\documentclass[letterpaper, 10 pt, journal, twoside]{IEEEtran}
\usepackage{glossaries}
\usepackage{subfigure}
\usepackage{caption}
\newtheorem{definition}{Definition}

\usepackage{graphicx}

\usepackage{amsmath,amssymb}
\usepackage{url}
\usepackage{hyperref}

\newcount\Comments  
\usepackage{color}
\definecolor{darkgreen}{rgb}{0,0.5,0}
\definecolor{purple}{rgb}{1,0,1}
\newcommand{\kibitz}[2]{\ifnum\Comments=1\textcolor{#1}{#2}\fi}

\newcommand{\yi}[1]  {\kibitz{black}   { #1}}
\newcommand{\xiaowei}[1]{\kibitz{blue}       {[Xiaowei: #1]}}

\newtheorem{remark}{\bf Remark}
\newcommand{\R}[1]{{\color{black}{#1}}}
\newcommand{\RR}[1]{{\color{black}{#1}}}
\newacronym{DRL}{DRL}{Deep Reinforcement Learning}
\newacronym{TD3}{TD3}{Twin-Delayed Deep Deterministic policy gradient}
\newacronym{UUV}{UUV}{Unmanned Underwater Vehicle}
\newacronym{OP}{OP}{Operational Profile}
\newacronym{AUV}{AUV}{Autonomous Underwater Vehicle}
\newacronym{MDP}{MDP}{Markov Decision Process}
\newacronym{DDPG}{DDPG}{Deep Deterministic Policy Gradient}
\newacronym{PID}{PID}{Proportional-Integral-Derivative}
\newacronym{PMC}{PMC}{Probabilistic Model Checking}
\newacronym{DTMC}{DTMC}{Discrete Time Markov Chain}
\newacronym{CTMC}{CTMC}{Continuous Time Markov Chain}
\newacronym{LTL}{LTL}{Linear Temporal Logic}
\newacronym{PCTL}{PCTL}{Probabilistic Computational Tree Logic}
\newacronym{RAS}{RAS}{Robotics and Autonomous Systems}
\newacronym{DNN}{DNN}{Deep Neural Networks}

\ifCLASSINFOpdf
\else
\fi
\begin{document}
%
\title{Reachability Verification Based Reliability Assessment for Deep Reinforcement Learning Controlled Robotics and Autonomous Systems}
%
%
%

\author{Yi Dong$^{1}$, Xingyu Zhao$^{2}$, Sen Wang$^{3}$, and Xiaowei Huang$^{1}$%
\thanks{Manuscript received: November, 22, 2023; Accepted January, 29, 2024.}
\thanks{This paper was recommended for publication by Editor Jaydev P. Desai upon evaluation of the Associate Editor and Reviewers' comments.
This work was supported by UK Dstl through Safety Argument for Learning-enabled Autonomous Underwater Vehicles and UK EPSRC through End-to-End Conceptual Guarding of Neural Architectures [EP/T026995/1].
This project has received funding from the European Union’s Horizon 2020 research and innovation programme under grant agreement No 956123.)} 
\thanks{$^{1}$Yi Dong and Xiaowei Huang are with Department of Computer Science, University of Liverpool, the UK
        {\tt\footnotesize \{yi.dong,xiaowei.huang\}@liverpool.ac.uk}}%
\thanks{$^{2}$Xingyu Zhao is with WMG, University of Warwick, Warwick, U.K.
        {\tt\small xingyu.zhao@warwick.ac.uk}}
\thanks{$^{3}$Sen Wang is with the Department of Electrical and Electronic Engineering, Imperial College London, London, U.K.
        {\tt\small sen.wang@imperial.ac.uk}}%
\thanks{Digital Object Identifier (DOI): see top of this page.}
}
%
%

\markboth{IEEE Robotics and Automation Letters. Preprint Version. Accepted January, 2024}
{Dong \MakeLowercase{\textit{et al.}}: Reachability Verification Based Reliability Assessment for DRL Controlled RAS} 

%



\maketitle

\begin{abstract}
Deep Reinforcement Learning (DRL) has achieved impressive performance in robotics and autonomous systems (RAS). A key challenge to its deployment in real-life operations is the presence of spuriously unsafe DRL policies. Unexplored states may lead the agent to make wrong decisions that could result in hazards, especially in applications where DRL-trained end-to-end controllers govern the behaviour of RAS.
This paper proposes a novel quantitative reliability assessment framework for DRL-controlled RAS, leveraging verification evidence generated from formal reliability analysis of neural networks. A two-level verification framework is introduced to check the safety property with respect to inaccurate observations that are due to, e.g., environmental noise and state changes. Reachability verification tools are leveraged locally to generate safety evidence of trajectories. In contrast, at the global level, we quantify the overall reliability as an aggregated metric of local safety evidence, corresponding to a set of distinct tasks and their occurrence probabilities. The effectiveness of the proposed verification framework is demonstrated and validated via experiments on real RAS.
\end{abstract}


%
\IEEEpeerreviewmaketitle

\section{Introduction}\label{sec_introduction}

\IEEEPARstart{D}{eep} Reinforcement Learning (DRL) has achieved impressive experimental results in video game playing, 
in which DRL agents are 
deployed under a trial-error-replay model. However, safety-critical applications normally are not able to trial and replay at will
in the real world, such as autonomous vehicles, power systems, and humanoid robots. 
A recent trend in autonomous navigation, including ground and underwater vehicles, is to use end-to-end controllers trained by reinforcement learning methods \cite{kiran2021deep}. 
For these applications that require a high level of safety integrity, deploying unverified DRL policies can lead to catastrophic consequences. 
In the meantime, \gls{DNN} in DRL algorithms 
are known to be unrobust to adversarial examples, 
i.e., the output of a DNN may be subject to dramatic change under a minor input disturbance. 
Such issues have motivated this work to formally analyse end-to-end DRL systems and assure their reliability before they are deployed.

Motivated by the question of whether local reachability verification can be extended to answer broader questions of system-level or model-wise reliability, we introduce a reliability metric based on the probabilistic notion of proportions of safety violations in the global input space.
Furthermore, most DRL algorithms are explored and learned in simulation environments for, e.g., efficiency and cost considerations.
However, the gap between simulated and real environments may lead to a violation of the safety property set in the simulation. 
Observation in the simulation environment is assumed to be an accurate and unbiased signal, which is an unacceptably strong assumption in a safety-critical context. 
\RR{To this end, we propose a novel two-level verification framework to assess the reliability of DRL algorithms: 
i) At the local level, tools for reachability analysis are used to generate evidence supporting safety verification, manifesting as interval-based\footnote{Unlike traditional point-based sampling methods, which verify only a single initial point and, consequently, a single trajectory at any given time, the interval-based approach enables the simultaneous verification of multiple initial points that fall within a specified interval.} results.
ii) At the global level, we borrow ideas from software reliability engineering to model the distribution of initial states as the OP\footnote{The \gls{OP} is a representation of a distinct set of tasks that a component or system performs \cite{musa1993operational}. These tasks may be influenced by user behaviour during interactions with the component or system. The profile also includes the probabilities associated with the occurrence of each task.} (representing the distribution of all possible operational scenarios), and then aggregate local safety evidence according to the OP to statistically estimate the overall reliability.}

The main contributions of this paper include:

\textit{a)} A two-level analysis framework is designed to assess the reliability of DRL algorithms with a focus on reachability verification. Reachability analysis formally verifies the safety of trajectories starting from an initial state at the local level, while overall reliability claims are supported statistically at the global level across initial states.
  
\textit{b)} \textcolor{black}{At the local level, the state-of-the-art reachability verification tool POLAR \cite{huang2021polar} is integrated and optimised for faster computation (about 19x faster than POLAR) and tighter bounded results. Meanwhile, OP is introduced and approximated to support global-level reliability claims.
}
   
\textit{c)} \R{A publicly accessible repository of our method with all source code, datasets, experiments and a real-world case study based on BlueRov2 Unmanned Underwater Vehicle (UUV) is provided.} 

{
\begin{table*}[h]
\centering
\caption{Connections between reliability assessment for traditional software, DL classifiers and DRL controllers.}
\label{table_compare}
\begin{tabular}{|c|c|c|c|}
\hline
                                                                    & Traditional safety critical  software                           & DL classifiers                                                                                                    & \textbf{DRL controllers}                                         \\ \hline
Metric                                                              & \begin{tabular}[c]{@{}c@{}}Probability of failure \\ per random demand ($\mathit{pfd}$)\end{tabular}                                  & \begin{tabular}[c]{@{}c@{}}Probability of misclassification\\  per random image ($pmi$)\end{tabular}                         & \begin{tabular}[c]{@{}c@{}}\textbf{Probability of crash per}\\  \textbf{random initial state}\end{tabular}                           \\ \hline
OP                                                                  & \begin{tabular}[c]{@{}c@{}}Distribution of\\ independent demands\end{tabular}                                                & \begin{tabular}[c]{@{}c@{}}Distribution of\\ independent images\end{tabular}                                                 & \begin{tabular}[c]{@{}c@{}}\textbf{Distribution of}\\ \textbf{independent initialisations}\end{tabular}                                          \\ \hline
\begin{tabular}[c]{@{}c@{}}Event of \\ interest\end{tabular}        & \begin{tabular}[c]{@{}c@{}}Failure that has \\ safety impact\end{tabular}                                                    & \begin{tabular}[c]{@{}c@{}}Misclassification that\\ has safety impact\end{tabular}                                           & \begin{tabular}[c]{@{}c@{}}\textbf{Failure that leads to} \\ \textbf{ hazards, e.g., crash}\end{tabular}                                                         \\ \hline
Partitions                                                          & \begin{tabular}[c]{@{}c@{}}Classes of input demands (``bins'') \\ based on  functionalities\end{tabular}                      & \begin{tabular}[c]{@{}c@{}}Norm-balls in the input space\\ with certain radius \\ (e.g., the $r$-separation distance \cite{yang2020closer})\end{tabular} & \begin{tabular}[c]{@{}c@{}}\textbf{Norm-balls in the input space with certain}\\ \textbf{radius representing a set of initial states}\\ (e.g., a specified bound of errors)\end{tabular}    \\ \hline
\begin{tabular}[c]{@{}c@{}}V\&V in \\ each partition\end{tabular}   & \begin{tabular}[c]{@{}c@{}}Estimation on conditional\\ $\mathit{pfd}$ of each bin (e.g., by stress\\ testing of a ``bin'')\end{tabular} & \begin{tabular}[c]{@{}c@{}}Local robustness estimation \\ within a given norm-ball\end{tabular}              & \begin{tabular}[c]{@{}c@{}}\textbf{Reachability verification}\\ (a ``strip'' of bounded trajectories in the input\\ space starting with a norm-ball)\end{tabular} \\ \hline
\begin{tabular}[c]{@{}c@{}}Oracle of \\ each partition\end{tabular} & \begin{tabular}[c]{@{}c@{}}By observation or given\\ by the specification\end{tabular}                                       & \begin{tabular}[c]{@{}c@{}}Label of the central-point\\ (seed) of the norm-ball\end{tabular}                                 & \begin{tabular}[c]{@{}c@{}}\textbf{Emptiness of the intersection of}\\ \textbf{the ``strip'' and predefined unsafe-areas}\end{tabular}                               \\ \hline
\end{tabular}
\end{table*}

}
%

\section{Related Work}\label{sec_relatedwork}
DRL has made significant strides in various domains, particularly in robotics and autonomous systems. As these technologies become more integrated into our daily lives and critical infrastructures, the emphasis on their safety and reliability becomes paramount. DRL algorithms train decision-making agents to select actions that maximize long-term performance. However, in many real-world scenarios, optimal performance alone isn't sufficient. Systems' robustness, stability, and safety must be rigorously ensured. To address these concerns, DRL verification and testing methods have emerged. These methods aim to guarantee a system's properties over an infinite set of inputs. Pattanaik \textit{et al.} first proposed adversarial attacks for DRL algorithms. By training DRL with engineered adversarial attacks, the performance is improved, leading to more robust DRL models \cite{pattanaik2017robust}. However, these adversarial training methods cannot guarantee safety during the training process. In response, Dalal \textit{et al.} designed a safety layer that solves an action correction formulation per state \cite{dalal2018safe}.
Felix \textit{et al.} utilized the Lyapunov function to determine the region of attraction for specific policies and applied statistical models to achieve high-performance DRL policies \cite{berkenkamp2019safe}. While these algorithms prioritize safety during training, the testing environment might differ. Hence, run-time monitors were introduced to ensure safety during operations. The shield structure prevents agents from making unsafe decisions by banning unsafe actions for each state \cite{alshiekh2018safe}.

The challenges of DRL verification are multifaceted. Unlike traditional Deep Learning (DL) verification, DRL involves sequential decision-making, where a Deep Neural Network (DNN) is invoked multiple times for each action decision \cite{van2017challenges}. This sequential nature, combined with the often stochastic environments in which DRL operates, introduces significant scalability challenges \cite{amir2021towards,eliyahu2021verifying}. For instance, in applications like self-driving cars, ensuring consistent actions for both perturbed and unperturbed states at every decision point is crucial. Various DRL verification approaches have been developed to address these challenges \cite{landers2023deep}. These can be broadly categorized into abstraction, constraint-based verification, reachability analysis, and model checking. Notably, our work aligns with the "Reachability Analysis" category, which includes prominent algorithms such as Verisig \cite{ivanov2019verisig,ivanov2021verisig}, Sherlock \cite{dutta2019reachability}, ReachNN \cite{fan2020reachnn}, POLAR \cite{huang2021polar}. However, a gap remains in our understanding of DRL safety. While existing methods can determine the presence of safety violations under extreme conditions, they often don't provide a comprehensive understanding of DRL policy safety, especially when violations are detected locally. This mirrors challenges faced in evaluating Deep Learning (DL) classifiers \cite{wangUAI21oxford}.

In conclusion, as DRL continues its transformative impact on robotics and autonomous systems, the quest for safety and reliability remains at the forefront. The field is evolving, with researchers continually proposing innovative techniques to bolster the safety, robustness, and reliability of DRL-driven systems. To better position the scope of our work, we summarise our approach for DRL, compared with reliability assessment methods designed for traditional software and DL classifiers, in Table \ref{table_compare}, where ours is highlighted in bold.

\section{PRELIMINARIES}\label{sec_preliminaries}

\subsection{Deep Reinforcement Learning} \label{sec:DRLpreliminary}

We use discounted infinite-horizon \gls{MDP} to model the interaction of an agent with the environment $E$. \RR{An MDP is a 5-tuple ${\cal M}^E=(\mathcal{S},\mathcal{A}, \mathcal{P}, \mathcal{R}, \gamma)$, where $\mathcal{S}$ is the state space, $\mathcal{A}$ is the action space, $\mathcal{P}(\textbf{x}'|\textbf{x},\textbf{a})$ is a probabilistic transition, $\mathcal{R}(\textbf{x},\textbf{a})\in {\mathbb R}_{\ge 0}$ is a reward function, $\gamma\in [0,1)$ is a discount factor, and $\textbf{a}\in\mathcal{A}$ represents a set of actions.} In our model, $\textbf{x}$ serves a dual purpose: it represents a state in the state space $\mathcal{S}$, and it is also used as the input to a policy neural network later in the process.  
\R{We consider DRL algorithms, such as deep deterministic policy gradient (DDPG)
, Twin Delayed DDPG (TD3)
, Proximal Policy Optimization (PPO)
}
, which return an optimal policy $\pi^*$, that includes a mapping $\mu: \mathcal{S} \rightarrow \mathcal{A}$ that maps from states to actions.

Based on $\mathcal{M}^E$, a policy $\pi$ induces a trajectory distribution $\rho^{\pi,E}(\zeta)$ where
$
    \zeta=(\textbf{x}_0,\textbf{a}_0,\textbf{x}_1,\textbf{a}_1,...)
$
denotes a random trajectory. The state-action value function of $\pi$ is defined as $Q^{\pi}(\textbf{x},\textbf{a})= \mathbb{E}_{\zeta\sim \rho^{\pi,E}}[\sum_{t=0}^{\infty}\gamma^t\mathcal{R}(\textbf{x}_t,\textbf{a}_t)]$
and the state value function of $\pi$ is
$
    V^\pi(\textbf{x})=Q^{\pi}(\textbf{x},\pi(\textbf{x})).
$ 

\subsection{Reachability Analysis and Verification}

Reachability analysis has been developed recently for the verification of the safety and robustness of DNNs \cite{ruan2018reachability,huang2019reachnn,huang2020survey,althoff2010reachability}. 
Adapted to the context described in Section~\ref{sec:DRLpreliminary}, reachability analysis determines the set of states that a system can reach, starting from a set of initial states and considering the interaction between the DRL policy and the MDP.
Safety verification, which is to determine whether a given DRL policy may lead to any unsafe state over an MDP, can be reduced to the reachability problem of whether an unsafe state is reachable.

\RR{\begin{remark}
    In different application contexts, the definition of a reachable ``safe set" may vary from case to case. Typically, it is defined based on safety analyses such as hazard identification for the given application \cite{qi2023stpa}. For instance, in applications related to battery systems, the ``safe set" could encompass safe charging and discharging intervals, while in power systems, it might involve maintaining system frequencies within safe operational bounds. In this paper, we have chosen to illustrate our approach using the example of a UUV, where the ``safe set" is represented by collision-free intervals.
\end{remark}}


In this paper, we verify a safety
property on a model-free DRL algorithm by computing its reachable set over a full trajectory. Similar to POLAR \cite{huang2021polar}, the following two mathematical principles are employed to calculate the reachable set: Taylor Arithmetic and Bernstein Polynomial. 

\subsubsection{Taylor Arithmetic}

Following \cite{makino2003taylor,huang2021polar,ivanov2021verisig}, any interval can be transferred into a Taylor model. A Taylor model is combined by a polynomial approximation $p$ and an interval error bound $I$: $TM = p(\textbf{x}) + I, \textbf{x}\in D$,
where $D$ is the input domain of the Taylor model. $I$ is the remainder of the Taylor model. Given two Taylor Models: $TM_1 = (p_1,I_1)$ and $TM_2 = (p_2,I_2)$, the addition and multiplication are computed as:
\small
\begin{align*}
TM_1 + TM_2 = & (p_1+p_2,I_1+I_2)\\
TM_1 \times TM_2 = &(p_1\times p_2 - r_k,I_1\times I_2+Int(p_1)\times I_2\\&+Int(p_2)\times I_1+Int(r_k))
\end{align*}\normalsize
where $Int(\cdot)$ is the interval of the polynomials, $k$ is the maximum order of the Taylor models, and $r_k$ is the truncated polynomial of the Taylor models. \R{To illustrate the calculation more clearly, we use a simple example to delve into the Taylor arithmetic \cite{makino2003taylor}. Consider two intervals $[0.4,1.1]$ and $[-1.2,1.3]$, we can transfer these to the TMs $(1-0.5x^2, [-0.1,0.1])$ and $(x+0.1x^4, [-0.2,0.2])$ over the domain $x\in[-1,1]$. 
    The order 4 TM for the sum is $(1+x-0.5x^2+0.1x^4, [-0.3,0.3])$, and the order 4 TM for the product is $(x-x^3+x^4, [-0.38,0.38])$.}

\subsubsection{Bernstein Polynomial}
There are several different activation functions in \gls{DNN}s, e.g., Sigmoid, ReLU, and Tanh. The Taylor model can only propagate with continuous activation functions, but DNNs may use piece-wise activation functions to fit the data, such as the ReLU function. Inspired by \cite{huang2021polar}, the Bernstein polynomial can be applied to calculate the Taylor models with discrete activation functions. 

To achieve over-approximation for safety, the conservative remainder for a Bernstein polynomial should be considered. \R{The conservative remainder of a Bernstein polynomial refers to the error-bounded portion that ensures the combined reachable set of the output polynomial and remainder is a safe over-approximation of the function it approximates \cite{huang2019reachnn}. }\RR{With the Taylor Arithmetic, the output Taylor Model of the activation function is ($p_{\sigma}, I_{\sigma}$ = $[-\epsilon, + \epsilon]$), where: }
\begin{equation}
\small
\label{Bern}
    p_\sigma = \sum_{i=0}^k\left(\sigma(a+\frac{b-a}{k}i)C^k_i\frac{(y_i-a)^k(b-y_i)^{k-i}}{(b-a)^k}\right). 
\end{equation}
and
\small
\begin{align}
\label{BernErr}
    \epsilon = \max_{i=0,\cdots,m}&\left(\left|p_\sigma\left(\frac{b-a}{m}(i+\frac{1}{2})+a\right)\right.\right.\nonumber\\&\left.\left.-\sigma\left(\frac{b-a}{m}(i+\frac{1}{2})+a\right)\right|+\frac{b-a}{m}\right)
\end{align}\normalsize
where $m$ is the sampling steps and $\epsilon$ is the conservative error bound of the Bernstein polynomial $I_\sigma = [-\epsilon,+\epsilon]$. \R{We remark that we use $\sigma$ represents the activation function of the neural networks}, $a, b$ are the infimum and supremum of the activation function input, $k$ is the maximum order of the Bernstein polynomial, \yi{$C_i^k$ is a binomial coefficient}, and $y$ is a sampled point between $a$ and $b$.

\subsection{Operational Profile based Reliability Assessment}
{In real-world scenarios, the agent can take different trajectories from different initial states under a given policy and an operational environment.
To support reliability claims for DRL in safety-critical applications, all possible trajectories need to be considered.
In real applications, the initial state (and its trajectory) of the agent usually obeys some distribution that can be approximated from data. 
}

The OP has been widely modelled in reliability assessment, which is applied to represent the occurrence probabilities of function calls and the distributions of parameter values \cite{koziolek2005operational}.
In other words, the OP is a Probability Density Function over the whole input domain $D$, and returns the probability of $\textbf{x}\in D$ being selected as the next random input. Later, we formally define the reliability metric of a DRL policy in a given environment, considering the OP.

\section{ALGORITHM DESIGN}\label{sec_alg}
{In this paper, we design a novel two-level framework for assessing the reliability of a DRL-controlled \gls{RAS} based on the reachability verification tools and statistical analysis technologies. 
At the local-level, safety verification is reduced to a reachability problem of checking whether an unsafe state is reachable from an initial state, while the global-level claims on reliability\footnote{\RR{In our framework, safety and reliability, while distinct concepts in system engineering, do not necessitate separate statistical reasoning. This is because, we consider safety-critical applications in which any failure has the potential to result in unsafe outcomes. Therefore, like \cite{kalra_driving_2016,zhao_assessing_2019}, reliability specifically pertains to the probabilities of occurrences of failures that have direct relevance to the system's safety, which unities the two terms in our context. While ``reachability'' is a computational problem that determines the corresponding output set (sometimes called a ``safe set" despite potential ambiguity, to maintain consistency with the initial paper \cite{huang2021polar}) given an input set, studied at the local level of our framework.}} can be obtained by \gls{OP}-based statistical methods.}


\subsection{Local-level Reachability Verification}\label{low_level}

\begin{figure}[h]
    \centering
    \includegraphics[width=\hsize]{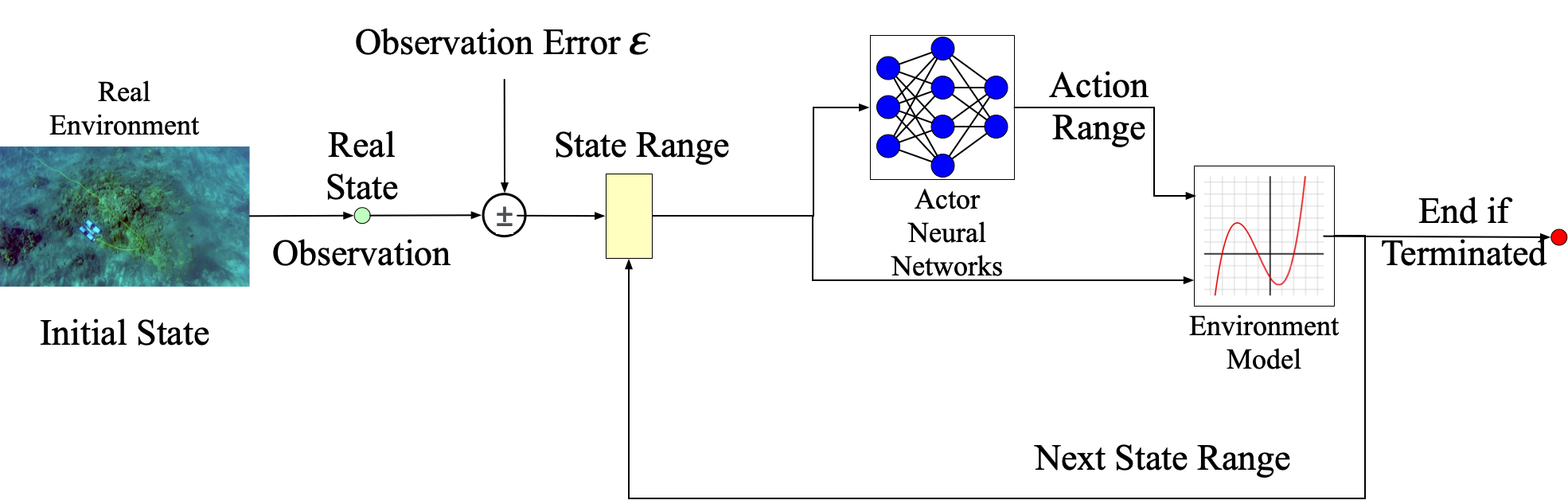}
    \caption{\R{Reachability verification framework.}}
    \label{framework}
\end{figure}

At the local level, we introduce an interval-based method to calculate the reachable set of a DRL policy. The DDPG algorithm is composed of an actor-network and a critic network. \R{We reduce the verification problem to a reachability problem, which is to calculate the reachable set of actions and states, starting from a set of initial states and considering the interaction between the DRL policy and the Environment, }
cf. Fig.~\ref{framework}.

\begin{figure}[h]
    \centering
     \includegraphics[width=0.9\hsize]{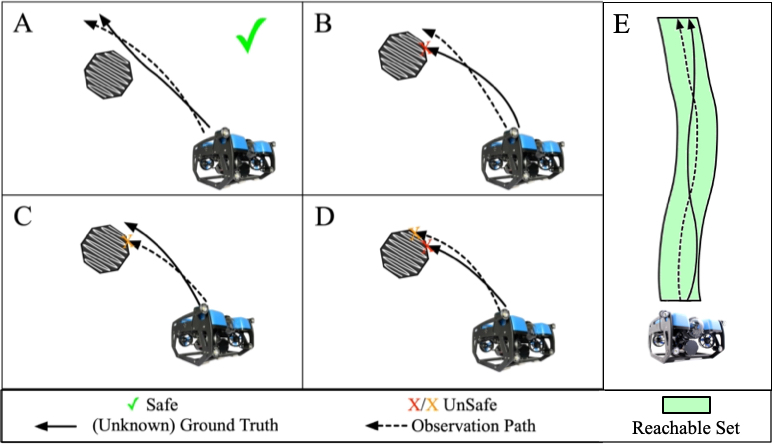}
     \caption{\RR{Different corner cases and illustration diagram (An example for UUV cases).}}
     \label{RAV}
\end{figure}

\R{In Fig. \ref{framework}, the green circle represents the "initial state", signifying a specific observation of the start point. The yellow square represents an initial set, which considered the observation errors, and therefore, this set encompasses the actual initial point. } The observation from the environment is normally noisy due to, e.g., inaccurate sensor signal and external disturbance. This noise could be ignored under safe conditions and in wide spaces but will cause safety issues in some corner cases, as illustrated in subfigures (A-D) of Fig. \ref{RAV}. 
It can be seen that the UUV is safe in case A because the real paths and observable paths are safe. 
In cases B, C, and D, the UUV is unsafe since at least one signal shows that a crash occurs. If the sensor did not observe the correct distance perception in a corner case, the decision of the current policy will 
lead to a crash with a high probability.
Consequently, these errors should also be considered in safety verification. Following the recommendation from \cite{agarwal2021deep}, the interval methods are 
suitable to deal with uncertain sensor noise. Based on the reachability verification tools, we use an interval to bound the real paths around the observable path. Subfigure (E) in Fig. \ref{RAV} shows that all the real paths are bounded in green intervals. 

Specifically, the robot's initial states are over-approximated with a range $\eta$ and all possible true states are theoretically bounded in this green interval. Then we have $\textbf{a}_t\in NN_{actor}(\textbf{x}),\ \textbf{x}\in\eta$, 
where $NN_{actor}$ is the actor neural network of DDPG.
Transferring the initial input domain into a Taylor model:
\begin{equation}
\small
    TM^0_i = \left(p_i(x_i),I_i\right), \, \, x_i\in\eta_i
\end{equation}
where $x_i$ is the $i$th state value of the observation. 
\R{\begin{remark}
    In the context of our work, the Taylor Model is innovatively adapted to represent input domains akin to norm balls. Within this representation, the polynomial component of the Taylor Model delineates the centre of the norm ball, while the remainder demarcates its radius.
\end{remark}}
The Taylor model can be directly propagated to the next layer.
\begin{equation}
\small
    tm^j_i = \sum_{k=1}^N\boldsymbol{w}_k\times TM^{j-1}_k + \boldsymbol{b}_i
\end{equation}
where $tm^j_i$ is the temporary Taylor model that has not passed the activation functions. $\boldsymbol{w}, \boldsymbol{b}$ are the weights and biases of $NN_{actor}$, respectively. $j, k$ are the indices of neural network layers and neurons, $N$ is the total number of neurons. 

\R{For the remaining layers of $NN_{actor}$, there are activation functions.}
However, Taylor arithmetic cannot process the interval input with the noncontinuous activation functions. Therefore, we employ the Bernstein polynomial to deal with the activation functions. Although the transformation will yield errors, we can summarise these errors into the remainder of the Taylor models by equations \eqref{Bern} and \eqref{BernErr}. 
\begin{equation}
\small
    TM_i^j = p_\sigma(\sum_{k=1}^N\boldsymbol{w}_k\times TM^{j-1}_k + \boldsymbol{b}_i) + Int(r_k) + I_\sigma,\ \ \ \ j>0
\end{equation}
where $\sigma$ represents the activation function of the neural networks. It is noticed that the maximum order of the Taylor models will be increased after the activation functions, and therefore we truncate the part polynomial $r_k$ of the output Taylor model to maintain the complexity of the Taylor models.
After several layers of propagation, the output Taylor model is 
\begin{equation}
\small
    TM^{out}_i = \sum_{k=1}^N\boldsymbol{w}_k\times TM^{out-1}_k + \boldsymbol{b}_i
\end{equation}
Here, the output Taylor model could be reversed to the interval by calculating the upper and lower bound of the Taylor model regarding the input domain $D$.


\subsection{POLAR Algorithm Optimisation}
In this paper, the POLAR algorithm is selected to test the DRL algorithm due to its advantages of tighter remainder bounds \cite{huang2021polar}. Different from other interval arithmetic methods, the POLAR algorithm generates better performance in the reachable set calculation, especially for DNNs. The deeper neural network layers cost POLAR a longer time to do Bernstein polynomial sampling. In consideration of the characteristics of the ReLU activation function, the propagation law of the Taylor models is separated into three parts, which depend on the interval ranges $TM\in[a,b]$. 

\begin{figure}
    \centering
    \includegraphics[width=\columnwidth]{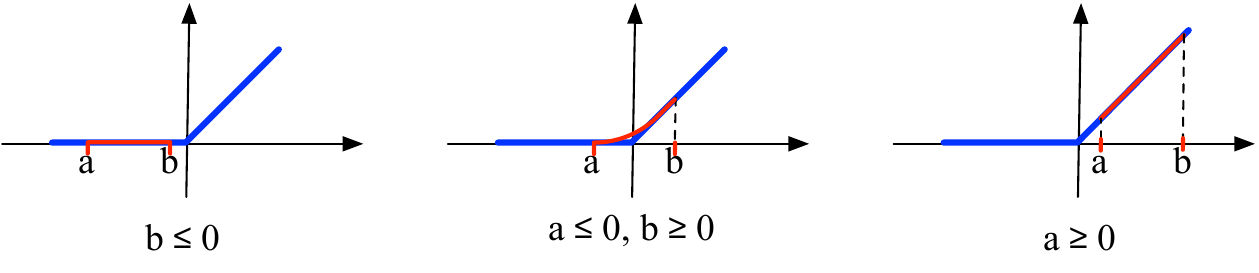}\
    \caption{\R{An Illustrative Framework of Equation \eqref{eq6}.}}
    \label{TMRELU}
\end{figure}

Furthermore, it is time-consuming to calculate the accurate bound of the high-order Taylor models. Similar to \cite{huang2021polar}, we calculate the conservative bound for the Taylor models with the Minkowski addition:
\begin{equation}
\small
    P \oplus Q = \{\textbf{p} + \textbf{q}|\textbf{p}\in P,\textbf{q}\in Q\}
\end{equation}
\R{Specifically, in our case, $P$ and $Q$ are two given sets that represent the range of the polynomial and the remainder interval. The variables $p$ and $q$ are elements within these sets $P$ and $Q$.} Consequently, to accelerate the testing speed and further enhance the testing accuracy of ReLU activation functions, we define the following propagation law for the Taylor models:
\begin{equation}
\footnotesize
\label{eq6}
TM^{o}=\left\{
\begin{aligned}
&0 , &if\ \ \ \  b\leq 0, \\
p_\sigma(TM^{i}) + &Int(r_k) + I_\sigma , &if\ \  a\leq0\ \ \&\ \  b\geq 0, \\
&TM^{i} , &if\ \ \ \  a\geq 0,
\end{aligned}
\right.
\end{equation}
where $TM^{i}$ and $TM^{o}$ are the Taylor models before and after the Relu activation function, respectively. The TM propagation law \eqref{eq6} is divided into three segments, as shown in Fig. \ref{TMRELU}. Due to the characteristics of the ReLU function, when $b$ is negative, the output of the function is consistently zero. Conversely, when $a$ is positive, the output TM of the activation function is equal to the input TM. If the input interval interacts with zero, we introduce the Bernstein polynomial to propagate the TM by Taylor arithmetic and approximate the output range.

\R{\begin{remark}
    Leveraging the capabilities of the Bernstein polynomial, our approach is applicable to all activation functions. In most cases, the majority of activation functions are ReLUs. Given this prevalence, we've optimized our method specifically for the characteristics of the ReLU function: we employ the Bernstein polynomial estimation only when the activation region encompasses zero. This decision stems from our empirical observations in real-world experiments, where most instances had the ReLU activation functions in an inactive state. By omitting calculations in these inactive scenarios, we save computational time and reduce potential errors. Furthermore, given our use of interval algorithms, we've incorporated an additional optimization: when all values within an interval are active, the output directly equates to the input, streamlining the process further.
\end{remark}}

{We enhance the POLAR algorithm by first splitting the polynomial fit on each neuron into three piecewise functions based on the characteristics of the ReLU function and then using the Bernstein polynomial to calculate the TMs if and only if the interval of TMs contains $0$. It can be seen that the proposed approach \eqref{eq6} is much faster than the original POLAR algorithm since the sampling steps of the Bernstein polynomial are excluded when the lower and upper bounds of TMs are positive or negative simultaneously. Nevertheless, the proposed method \eqref{eq6} can get the results without error (when the interval is positive-definite) or even eradicate the errors passed from the previous layer to zero (when the interval is negative-definite). Therefore, this optimisation step can reduce the computation time and increase accuracy.}


\subsection{Global-level Reliability Assessment}\label{high_level}
\R{The execution of the DRL policy for controlling the RAS in an environment leads to a trajectory distribution} (modelled as a Discrete-Time Markov Chain, discussed later), where the uncertainty (modelled with probability distribution) is from the environment\footnote{For simplicity and without loss of generality, we assume DRL policy is deterministic, but our method can be adapted for probabilistic policies.}. Formally, given an environment $E$, a policy $\pi$, and an initial state $\textbf{x}_0$, we can construct a model ${\cal M}^E(\pi,\textbf{x}_0)$ representing the probability distribution of a set of trajectories. Assume that we have a verification technique $g$, as discussed in Sections~\ref{low_level}.

\begin{definition}
The verification problem is, given a constructed model ${\cal M}^E(\pi,\textbf{x}_0)$ and a verification tool $g$, to determine whether the model is safe with respect to a certain property $\phi$, written as ${\cal M}^E(\pi,\textbf{x}_0)\models^{g} \phi$. We may omit the superscript $g$ and write ${\cal M}^E(\pi,\textbf{x}_0)\models \phi$ if it is clear from the context.  We can also assume that $g$ returns a probability value -- a Boolean answer can be converted into a Dirac probability. Then, the verification problem is to compute $Pr({\cal M}^E(\pi,\textbf{x}_0), \phi)$, i.e., the probability of safety.
\end{definition}

In the following, we discuss how the above verification problem may contribute to the computation of reliability.  Similar to \cite{dong_reliability_2022,zhao_assessing_2021}, we partition the space of initial states into $d$ sets, each of which is represented as a constraint $\mathcal{C}_i$, for $i=1..d$. 
Based on these, we can estimate the reliability (defined as the probability of failure in satisfying $\phi$ with the policy $\pi$ in the environment $E$) as
\begin{equation}\label{pdfeq}
\small
   \text{Reliability}(E,\pi,\phi) = \sum_{i=1}^m G_{\theta}(\mathcal{C}_i)(1-Pr({\cal M}^E(\pi,\textbf{x}_{\mathcal{C}_i}), \phi))
\end{equation}
\R{where $m$ represents the sample size in the summation,} $G_{\theta}(\mathcal{C}_i)$ returns the probability density of the partition $i$ that is  represented as the constraint $\mathcal{C}_i$, $\textbf{x}_{\mathcal{C}_i}$ denotes the central point (i.e., a representative) of $\mathcal{C}_i$, and $1-Pr({\cal M}^E(\pi,\textbf{x}_{\mathcal{C}_i}), \phi)$ returns the failure rate of the DRL agent $\pi$ working on inputs satisfying the constraint $\mathcal{C}_i$ under the environment $E$. Note that $G_{\theta}$ can be estimated in the same way as the data distribution in 
\cite{dong_reliability_2022,zhao_assessing_2021}. 

\section{EXPERIMENT RESULTS}\label{sec_sim}


\R{This section outlines three experimental cases to showcase the performance of the proposed method. The first case provides local-level verification to validate the algorithm's effectiveness. The second case compares the original POLAR algorithm and the Monte-Carlo sampling method, illustrating the algorithm's conservativeness through global-level reliability analysis. The final case applies the algorithm to real underwater robots, visually demonstrating safe and unsafe reachability verification, underscoring the algorithm's effectiveness and compatibility.}






\subsection{Experiment Setup}

\begin{figure}[htbp]
\centerline{\includegraphics[width=\hsize]{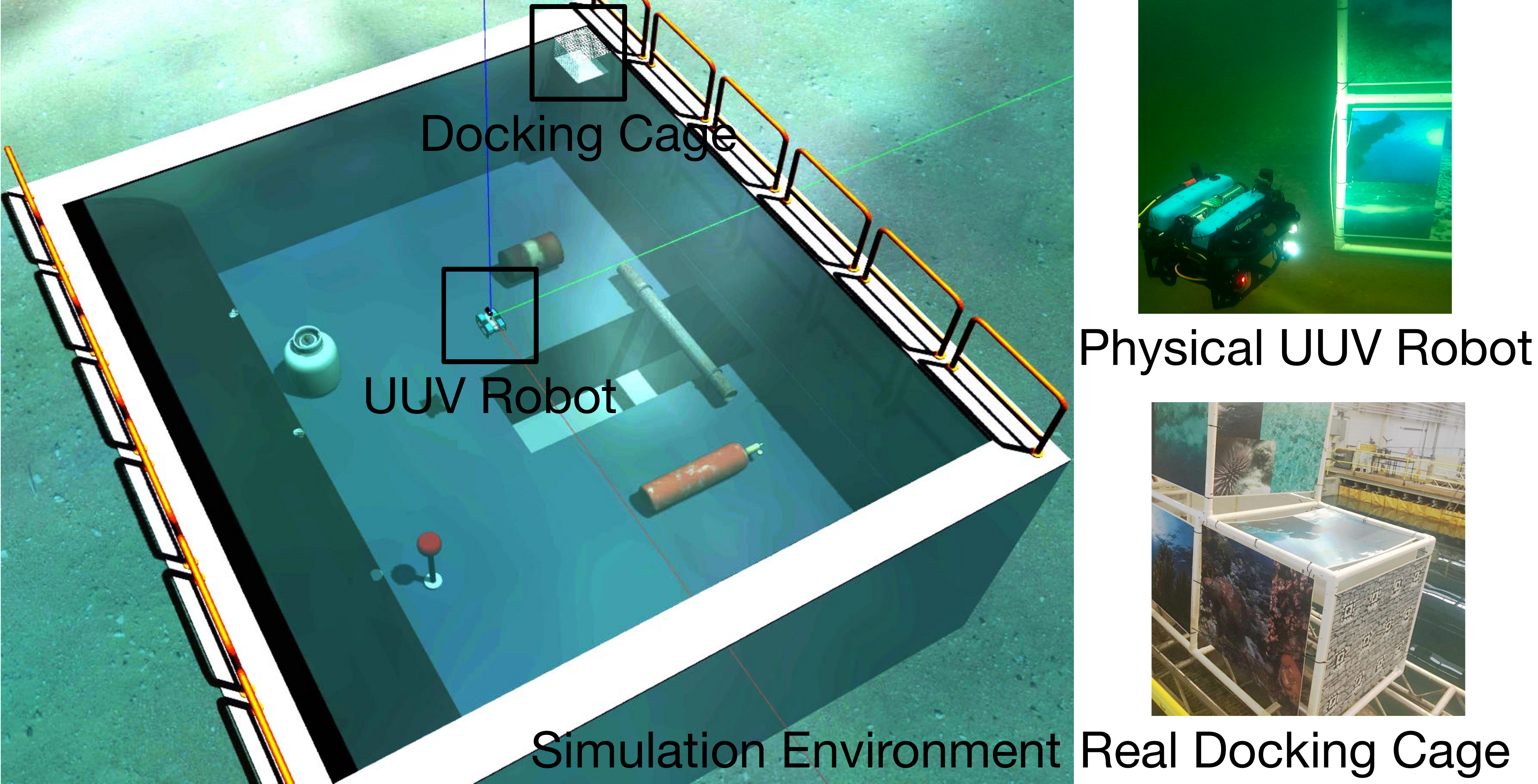}}
\caption{\RR{Simulation and physical experiment environments.}}
\label{ros+gazebo}
\end{figure}

In this experiment, we consider
a DRL-driven robot
that
navigates, and avoids collisions in a mission of automatically docking into a specified cage. 
We chose the BlueROV2 UUV\footnote{All source code, DRL models, datasets, and experiment results are available at solitude website \url{https://github.com/Solitude-SAMR}} as our subject for both simulated and physical experiments. In the simulated environment, we primarily utilized ROS and Gazebo, as depicted on the left side of the accompanying Fig. \ref{ros+gazebo}. For the physical experiment, the training process was conducted in a real water tank equipped with a docking cage, as shown on the right side of Fig. \ref{ros+gazebo}.

\subsection{Case 1: Local-level Verification}

\begin{figure}[htbp]
    \centering
     \includegraphics[width=0.8\hsize]{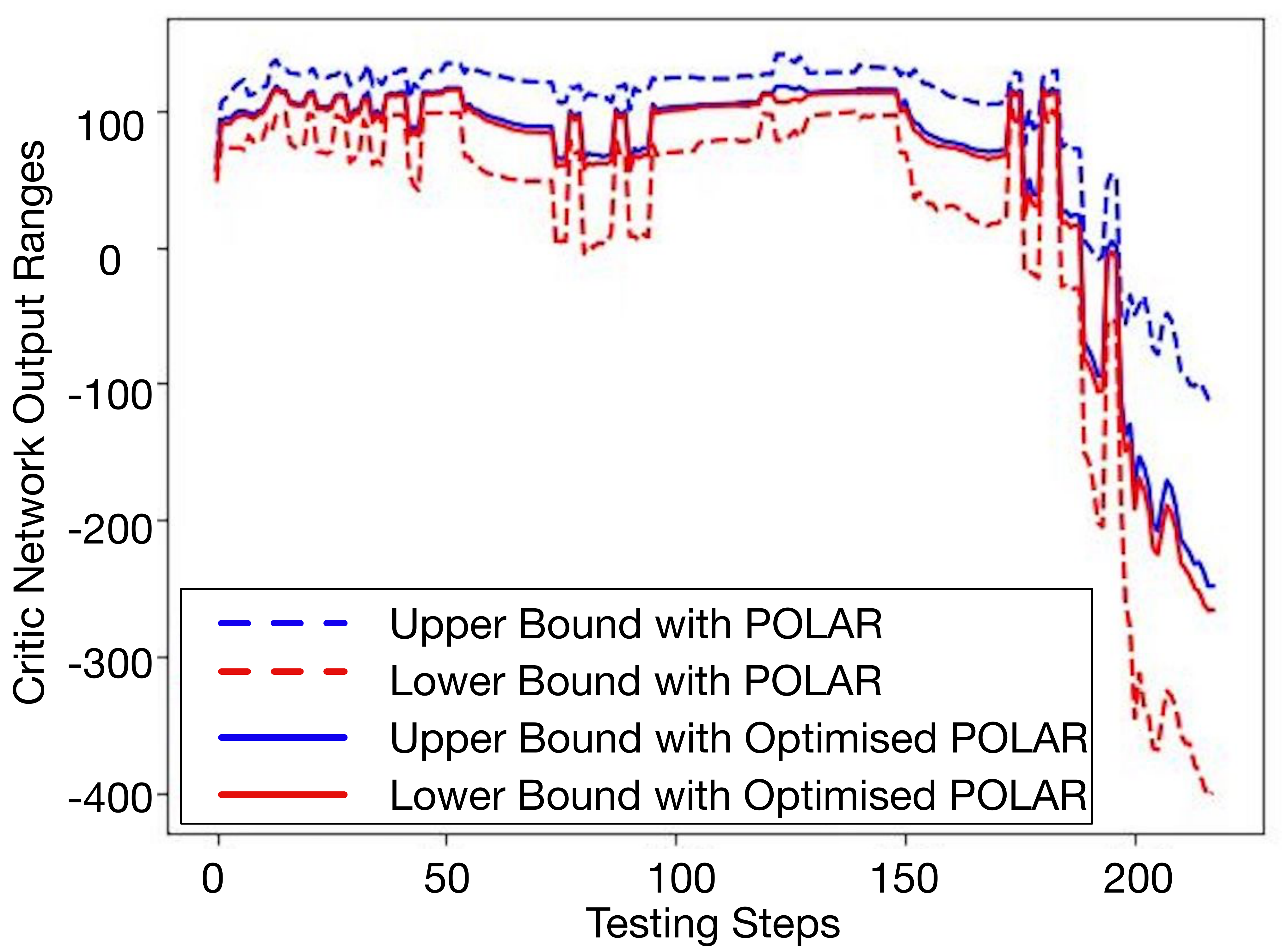}
     \caption{Comparisons between the original POLAR and our optimised POLAR---the latter yields tighter bounds. \vspace{-6pt}}
     \label{comparePOLAR1}
\end{figure}

\R{In this case, we illustrate local-level reachability verification using an optimized version of the POLAR algorithm, adapted for our TD3 policy's unique characteristics, enhancing precision in verifying reachability within specific environments. Our framework is versatile, allowing the incorporation of any reachability verification algorithm, with POLAR chosen for its current effectiveness in the field. We compared the over-approximated range of our method to the original POLAR algorithm, conducting experiments in a uniform Python environment, maintaining identical initial states, system dynamics, and neural network models for fairness. The original POLAR algorithm showed less tight bounds, particularly after 50 iterative steps, prompting updates to the input state interval at each step. Comparative performances and output $Q$ ranges are visually represented in Fig.~\ref{comparePOLAR1}.}

\yi{Upon examining Fig. \ref{comparePOLAR1}, it becomes evident that our optimization of the original POLAR algorithm significantly enhances its performance. While both algorithms are capable of obtaining the reachable set of the DNNs, our proposed approach consistently yields tighter bounds. This results in more accurate over-approximations, underscoring the superior performance and precision of our optimized algorithm. To underscore the superior performance of our algorithm in comparison to the POLAR algorithm, we carried out a comparative testing experiment under identical test conditions, utilizing a single CPU core and 500 samples. The results of this experiment are detailed in Table. \ref{timetable}.}

\begin{table}[htbp]
\centering
\begin{tabular}{ccc}
\cline{1-3}
     & POLAR & Our Method  \\ \cline{1-3}
Mean &  128.53     & 6.70732    \\
Var  &  6.8366     & 0.90574   \\
Max  &  134.68     & 11.57376  \\
Min  &  124.45     & 5.49535   \\ \cline{1-3}
\end{tabular}
\caption{Computing time comparison (results in seconds).}
\label{timetable}
\end{table}

A glance at the table reveals that our algorithm achieves a speed increase of approximately 94\% compared to the POLAR algorithm. This substantial improvement can be attributed to the inefficiencies inherent in the POLAR method. Specifically, POLAR conducts unnecessary Bernstein sampling for a large number of neurons that are either entirely inactive or fully activated during neural network inference. In contrast, our method introduces a more nuanced approach. We meticulously categorize different activation states and eliminate a significant amount of redundant sampling. This strategic reduction in unnecessary computations results in a considerable saving of testing time, thereby enhancing the overall efficiency and performance of our algorithm.

\subsection{Case 2: Global level Reliability Assessment}

\begin{figure}[htbp]
    \centering
     \includegraphics[width=0.8\hsize]{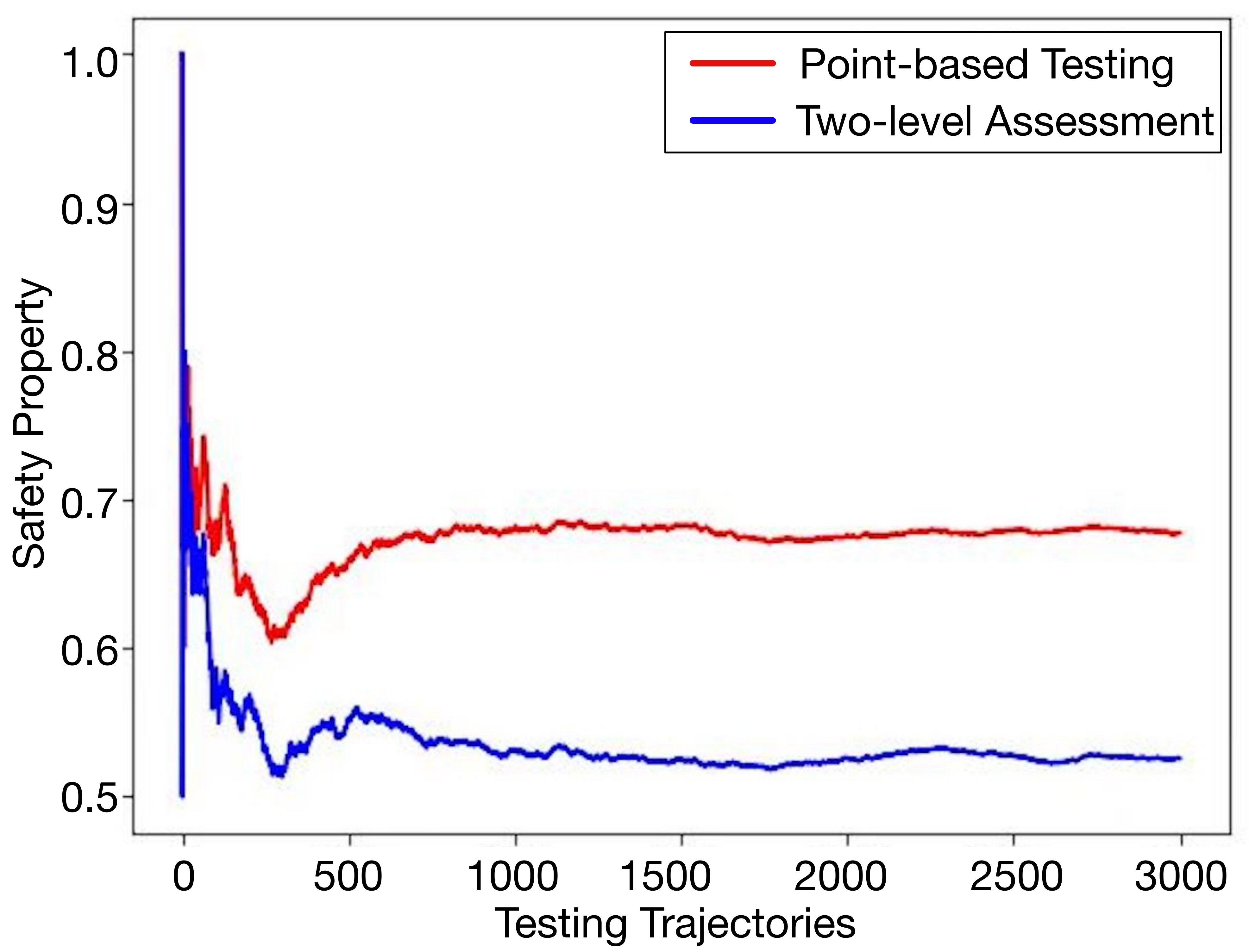}
     \caption{Comparisons between Monte-Carlo sampling and our two-level interval-based assessments---the latter yields more conservative results. \vspace{-6pt}}
     \label{comparePOLAR2}
\end{figure}

After local verification, we assessed the reliability of the DRL policy to validate our framework's efficacy. For a given DRL policy and initial spaces, we estimated system reliability using equation \eqref{pdfeq}. We collected data from a real-world water tank environment and tested it across 500 different initial states. Subsequently, we calculated the reachable set and compared the reliability of the UUV system using our method and the traditional Monte-Carlo method, with the results presented in Fig.~\ref{comparePOLAR2}.

\R{Fig.~\ref{comparePOLAR2} illustrates that the proposed interval-based approach is more conservative compared to the Monte-Carlo sampling method, as it considers the entire environment, providing an over-approximation of all possible points. We omitted the assessment results with the original POLAR algorithm due to its rapid error accumulation, making almost all initial intervals appear unsafe and rendering the establishment of reliable results impossible. This underscores the significance of our optimisations to the POLAR algorithm for balanced and practical system reliability assessment.

Additionally, Fig.~\ref{comparePOLAR2} reveals a convergence in the predicted reliability of the UUV system as the number of sampled initial states increases. The observed low reliability is attributed to the omission of safety factors during the DRL model training. The model's goal is to dock the UUV efficiently, prioritising reward and speed over safety, impacting the system's reliability assessment.}

\subsection{Case 3: Real Robot Experiment}
\begin{figure}[htbp]
    \centering
     \includegraphics[width=0.8\hsize,height=0.5\linewidth]{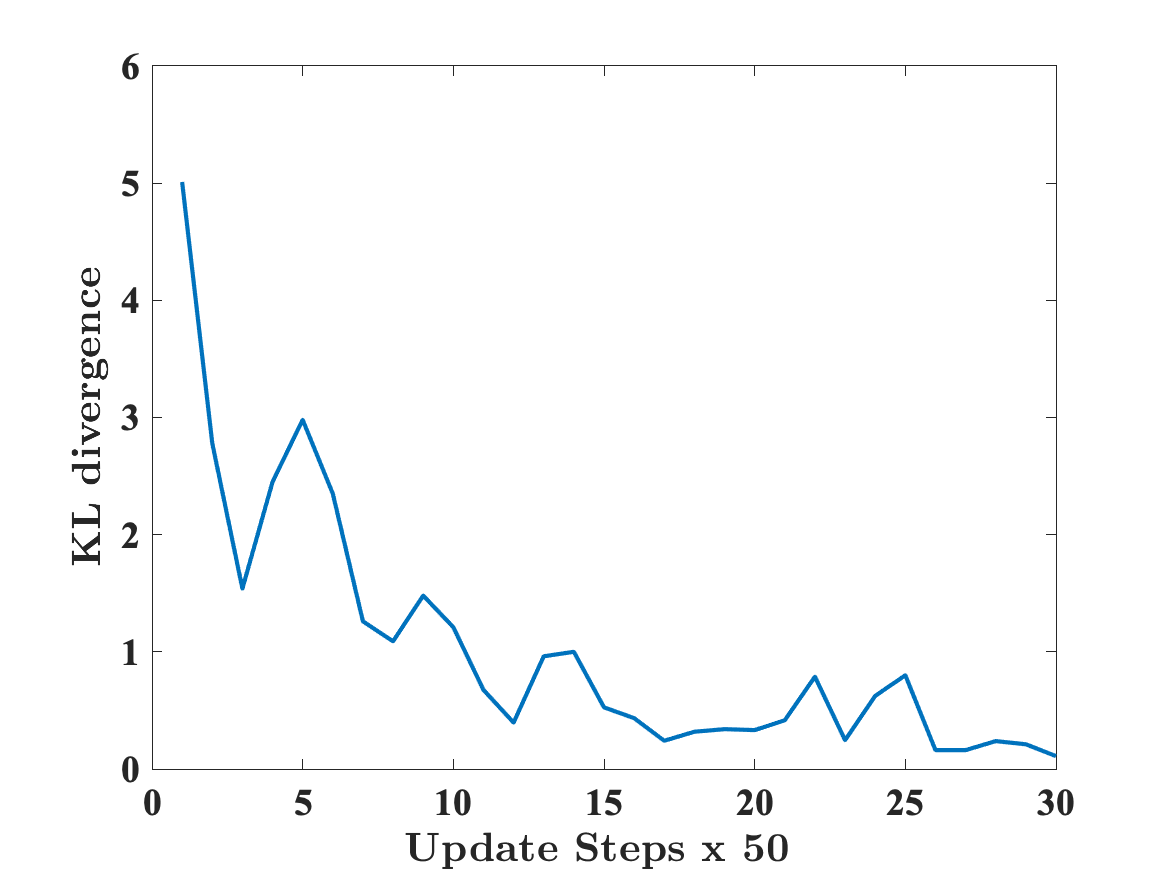}
     \caption{Convergence of the OP.}
     \label{fig:KL}
\end{figure}

\begin{figure}[htbp]
    \centering
     \includegraphics[width=\hsize]{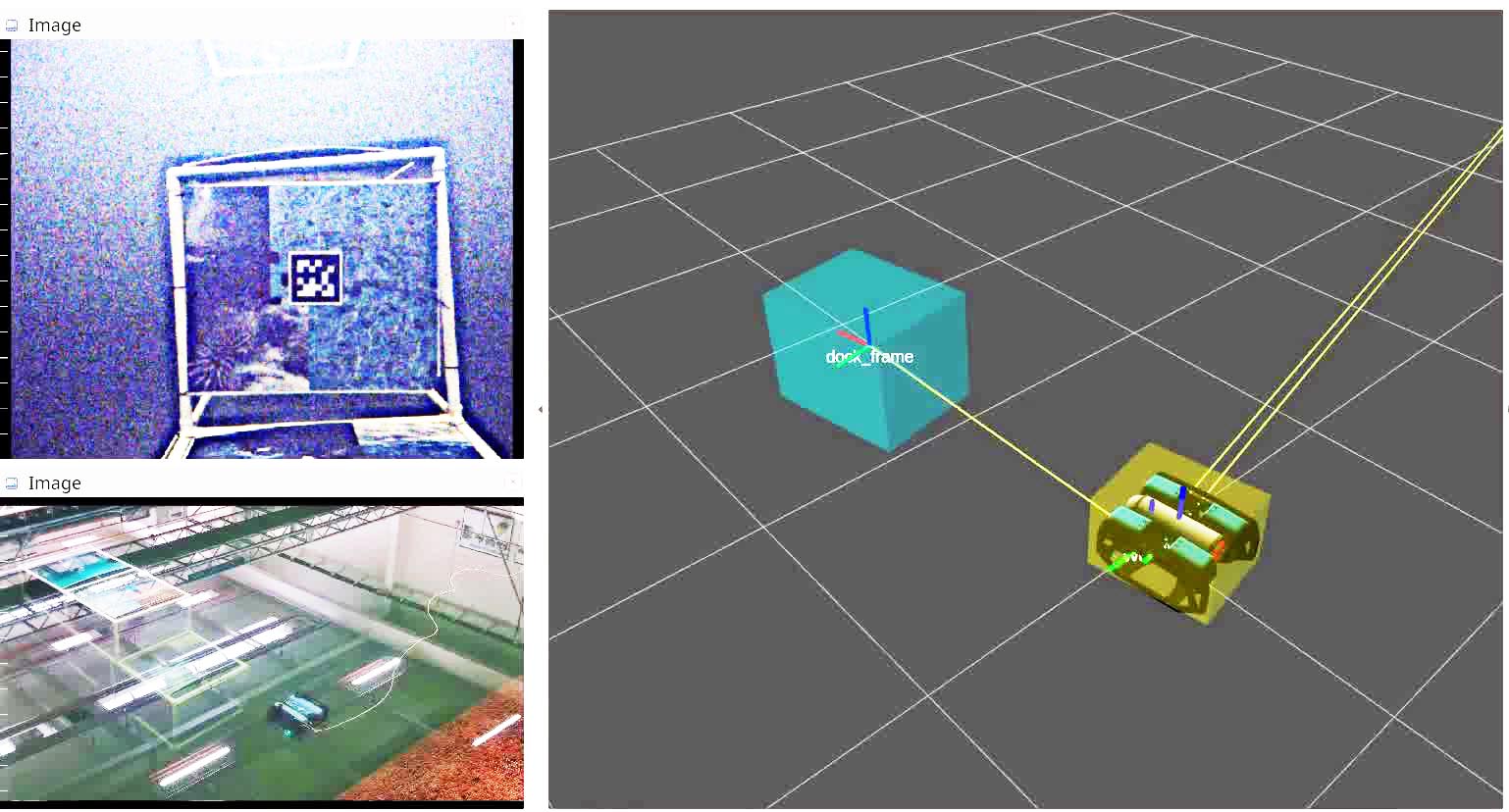}
     \caption{A case for verified safe docking mission.}
     \label{fig:safe}
\end{figure}

In this experiment, we began by randomly initialising the UUV from various starting points, which naturally generated the \gls{OP} within our reliability assessment framework. First, we employed Kullback-Leibler (KL) divergence as a metric to quantitise the convergence of OP. We calculate the KL divergence between distributions obtained at different iterations, specifically at intervals of 50 samples. \RR{The results of this process are visually represented in Fig. \ref{fig:KL}, which shows that the KL divergence values approach zero as the number of samples increases.}
This indicates that our sampling method is effective in providing a reliable estimation of safety across the entire input space. In an ideal scenario, we would be able to accurately estimate the safety of the model if all verified initial intervals could encompass the entire initial space. However, due to constraints such as computing power and scalability issues, we adopted a sampling method to estimate the safety of the entire input space in this study. Despite the limitations, our approach offers a practical and efficient solution for assessing the safety of the model in a comprehensive manner.
It is noticed that our research is intrinsically tied to safety considerations, and therefore it would be both impractical and hazardous to conduct experiments that involve crashing actual UUVs into docking cages. To mitigate this, we utilized real-world data to create a virtual environment where we could conduct safety-focused studies. Upon the successful completion of these simulation analyses and the verification of safe trajectories, we felt assured enough to apply our algorithm to real UUV BlueRov2, albeit initiating from verified safe points. The real-world tests served to validate our findings and are visualized through RViz. \RR{A verified safe docking process\footnote{Due to the page limit, the whole docking process can be viewed on \url{https://youtu.be/WRp1mm50En4}} is shown in Fig. \ref{fig:safe}.}

Fig. \ref{fig:safe} contains three key images that provide a comprehensive view of our real-world experiment. In the top-left corner, we can see real-time footage captured by the UUV's onboard front camera. This footage features a QR code, which serves as a crucial element for locating the position of the docking cage within the environment. Meanwhile, the image in the bottom-left corner offers an aerial view of our tank, captured using an overhead camera. This perspective provides a broader context for the experiment, showing the UUV, the docking cage, and other relevant elements within the lab. 
On the right side of the figure, the blue 3D box represents the virtual docking cage, while the yellow-green 3D box indicates the reachable region of UUV's next move. This reachable region signifies that regardless of the action output by the policy, the UUV will appear within this area at the next time step. In this scenario, the reachable region does not collide with the docking cage throughout the UUV's movement process. Therefore, we claim that the UUV, under the control of this policy, is safe for this initial range.

\begin{figure}[htbp]
    \centering
     \includegraphics[width=\hsize]{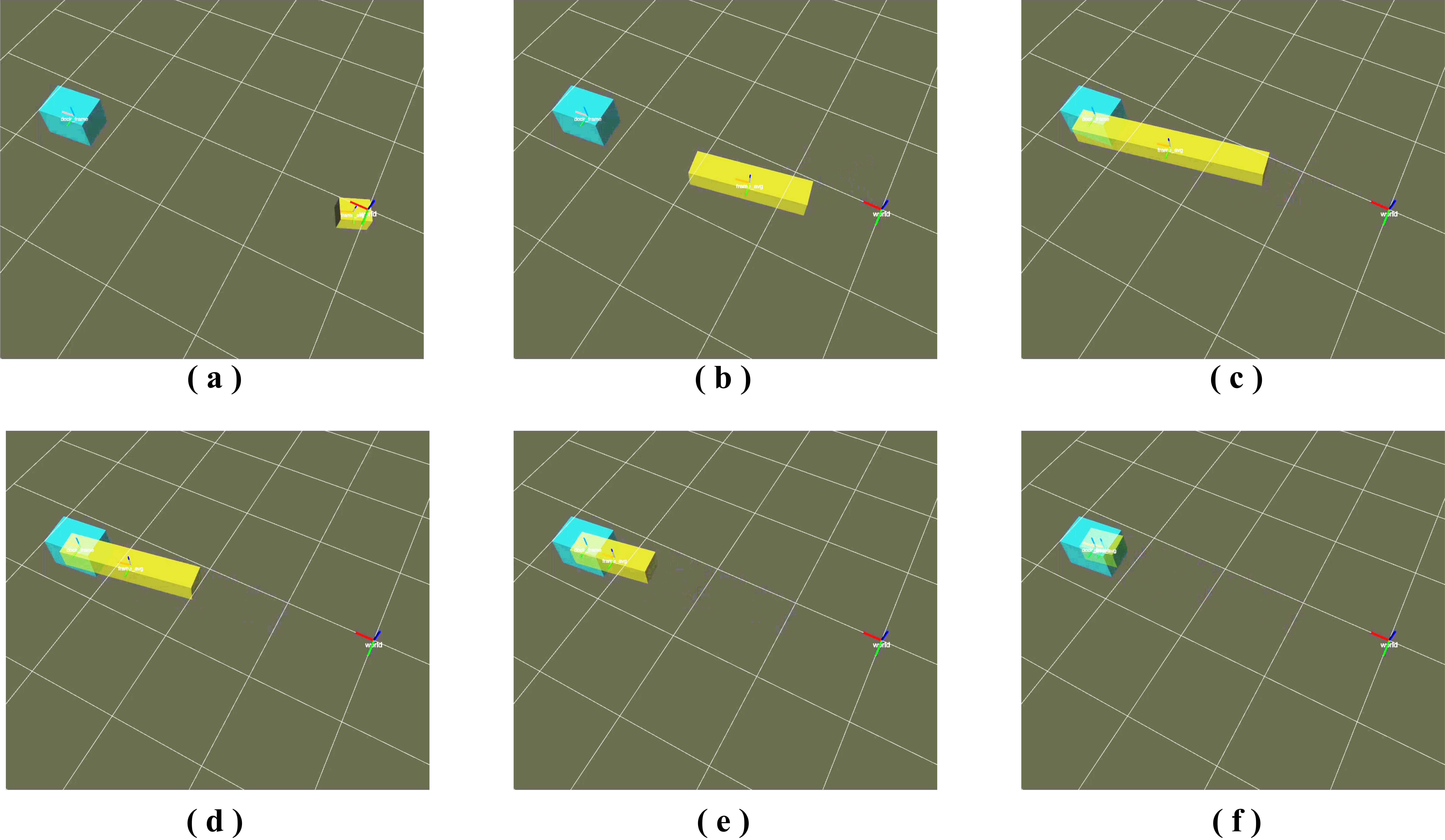}
     \caption{Verification results for an unsafe docking case.}
     \label{fig:unsafe}
\end{figure}

\yi{From the observation of the docking process, we can deduce that for a specific environment and policy, there are certain initial regions where the accumulated errors of the reachable region do not proliferate swiftly. This translates to the corresponding reachable region maintaining a controlled oscillation between expansion and contraction, providing a stable operational range for UUV. However, there are exceptions to this observation, particularly in what we term as $corner$ $cases$, as shown in Fig. \ref{fig:unsafe}.}

\yi{As presented in Fig. \ref{fig:unsafe}, these corner cases are characterised by a rapid expansion of the reachable region, triggered by the cumulative effect of errors. This rapid expansion occurs even when the current position of the UUV is at the side of the reachable region. A critical aspect to note here is the potential risk of collision between the expanding reachable region and the docking cage, as shown in sub-figures (c) to (e). Despite the UUV not being at the edge of the reachable region, the mere possibility of a collision makes these starting points unsafe. This potential hazard necessitates a more conservative approach in our reliability assessment, which is why our conclusions lean towards caution when compared to the results derived from Monte Carlo sampling methods.}

\section{Conclusion}\label{sec_con}
This paper studies the probability of failures that can cause hazards in DRL-controlled RAS.
A two-level reliability assessment framework is proposed, using reachability verifications at the local level and statistically supporting probabilistic reliability claims based on the operational profile at the global level.
An optimisation of the local-level reachability analysis algorithm is applied to enhance the verification speed and accuracy. 
The results in the simulation and the real world manifest the effectiveness of the proposed framework.
However, it is important to note that the two-level structure in local verification requires complete information, such as the physical environmental system model and environmental dynamics, which could be a limitation in certain scenarios. Additionally, while our method accelerates the existing algorithms, scalability remains a challenge. Future work could focus on developing approximation methods to mitigate the need for complete information and further algorithmic optimization to address scalability issues.

\end{document}